\pdfoutput=1

\documentclass[11pt]{article}
\usepackage{booktabs}
\usepackage[]{ACL2023}
\usepackage[title]{appendix}
\usepackage{times}
\usepackage{latexsym}
\usepackage{graphicx}

\usepackage[T1]{fontenc}

\usepackage[utf8]{inputenc}

\usepackage{microtype}

\usepackage{inconsolata}

\title{Turkish Native Language Identification V2: \\  L1 Influence of Arabic, Persian, and Albanian}

\author{Ahmet Yavuz Uluslu \\
  University of Zurich \\
  \texttt{ahmetyavuz.uluslu@uzh.ch} \\\And
  Gerold Schneider \\
  University of Zurich \\
  \texttt{gschneid@cl.uzh.ch} \\}

\begin{document}
\maketitle
\begin{abstract}
This paper presents the first application of Native Language Identification (NLI) for the Turkish language. NLI is the task of automatically identifying an individual's native language (L1) based on their writing or speech in a non-native language (L2). While most NLI research has focused on L2 English, our study extends this scope to L2 Turkish by analyzing a corpus of texts written by native speakers of Albanian, Arabic and Persian. We leverage a cleaned version of the Turkish Learner Corpus and demonstrate the effectiveness of syntactic features, comparing a structural Part-of-Speech n-gram model to a hybrid model that retains function words. Our models achieve promising results, and we analyze the most predictive features to reveal L1-specific transfer effects. We make our data and code publicly available for further study.\footnote{\href{https://github.com/projectauch/turkish-nli}{https://github.com/projectauch/turkish-nli}}
\end{abstract}

\section{Introduction}

Native Language Identification (NLI) is the task of automatically identifying the native language (L1) of an individual based on their linguistic productions in another language (L2). The underlying hypothesis is that the L1 influences learners’ second language writing as a result of the language transfer effect \citep{yu2016new}. It is used for a variety of purposes including forensics applications in cybercrime \citep{perkins2021application} and second language acquisition \citep{swanson2014data}. 

Research in NLI is mainly conducted with learner corpora, which comprise collections of writings by individuals learning a new language. These writings are annotated with metadata such as the author's native language (L1) or their fluency level. Recent NLI studies on languages other than English include Portuguese \citep{malmasi2018portuguese}, Arabic \citep{malmasi2014arabic}, and Chinese \citep{malmasi2014chinese}. The learner corpus is the backbone of NLI research, which means that extending research to a novel language depends on acquiring the appropriate learner corpora for that language. In the past, studies have focused on L2 English because of the prominence of this language in language research and the relatively large amount of data available. To the best of our knowledge, this study presents the first comprehensive NLI experiments on L2 Turkish. We employ the recently constructed Turkish Learner Corpus (TLC) \citep{anna2022error} and investigate widely used syntactic features for NLI. The remainder of this paper is organised as follows: Section 2 reviews related work, Section 3 and 4 describe our data and methodology, Section 5 presents the results and our discussion, and Section 6 concludes the paper.

\section{Related Work}
NLI is typically modeled as a supervised multi-class classification task. In this experimental design, the individual writings of learners are used to train a model while the author’s L1 information serves as class labels. A variety of feature types at the syntactic and lexical levels have been studied to capture distinct characteristics of the language interference phenomenon: spelling errors, word and lemma n-grams, dependency parsing, and morphosyntax. A more detailed review of linguistic features can be found in two shared task reports on the NLI task organised in 2013 and 2017 \citep{tetreault2013report, malmasi2017report}.

In recent years, there has been increased experimentation with deep learning methods, including pre-trained encoder models such as BERT \citep{steinbakken2020native} and decoder models such as GPT-2 \citep{lotfi2020deep}. While these models outperformed the state-of-the-art performance achieved by feature-based stacked classifiers \citep{malmasi2018native}, questions about their interpretability, inherent biases, and practical shortcomings in industrial applications remain unexplored. Traditional methods based on hand-crafted features continue to be preferred in many implementations due to their simplicity in training and resource efficiency. Within this context, \citet{uluslu2022scaling} approached the NLI scalability problem in the context of cybercrime through model compression and inference speed.

\section{Data}
In this study, we use data from the TLC \citep{anna2022error}. TLC is a learner corpus composed of the writings of learners of Turkish. These texts are essays written as part of a test of Turkish as a secondary language. While all learners in this study had completed a B1 level Turkish course, we observed a lack of standardized CEFR-based proficiency assessment in the material. Each text includes additional metadata such as the nationality of the author and the genre of the text. The corpus also includes error codes and corrections, although we do not make use of this information. In the second version of the NLI corpus, we removed learner texts from Azerbaijan (due to its close similarity to Turkish) and Afghanistan (due to the difficulty in determining a specific L1 among Pashto, Dari, and Persian).

We used a subset of the dataset containing texts for four L1 groups: Arabic (Syria), Albanian (Albania and Kosovo) and Persian (Iran). We additionally exclude individuals with Turkish origins from the Persian sample due to the prevalence of bilingualism in the Turcophone regions of Iran. The selection of these three languages is motivated by significant immigration trends in Turkey. This demographic context creates a practical need for NLI applications in areas such as digital forensics and Turkish language education. We limit our study to the genre of essays. The other genres (letters and petitions) in the corpus are unbalanced and scarce, which may introduce linguistic biases across different registers. We also do not attempt to adjust our dataset based on the writing prompts because they were unbalanced across languages. To normalize for document length, a known confounding variable, we chunked all texts for a given L1 into segments of approximately 200 tokens. This process ensures that each document is of a comparable length and represents a mix of authors and topics from that L1. The composition of our data is shown in Table~\ref{table:language_data_mean_sd}.

\begin{table}[h!]
\centering
\begin{tabular}{lccc}
\toprule 
\textbf{L1} & \textbf{$D$} & \textbf{Tokens} & \textbf{Mean $\pm$ SD} \\ 
\midrule 
Arabic & 76 & 12546 & 197.6 $\pm$ 67.3 \\
Albanian & 73 & 10224 & 140.1 $\pm$ 91.7 \\
Persian & 79 & 11712 & 148.8 $\pm$ 64.2 \\
\bottomrule 
\end{tabular}
\caption{Corpus statistics for the three L1 backgrounds. $D$ indicates the number of documents. The table also shows the total token count, and the mean and standard deviation (SD) of the document length (tokens per document).}
\label{table:language_data_mean_sd}
\end{table}
\section{Methodology}

\subsection{Classifier}
For this NLI task, we employ a standard supervised multi-class classification approach. Following previous research \citep{gebre2013improving}, we use a linear Support Vector Machine (SVM) classifier. The input features are n-grams weighted with the Term Frequency-Inverse Document Frequency (TF-IDF) scheme, which our preliminary experiments showed outperformed simple relative frequencies. To optimize the model, we performed a grid search for the regularization parameter C over the range [$10^{-3}$, $10^{-1}$] and set \texttt{max\_iter=2000} to ensure convergence. The model's performance plateaued at the upper end of this range, so we selected $C=1.0$ for our final experiments.

\subsection{Evaluation}
Following the previous NLI studies, we present our findings using classification accuracy through 5-fold cross-validation (5-FCV), which has become the standard for NLI result reporting in recent years. Our cross-validation approach is randomised and stratified, aiming to maintain consistent class proportions across partitions. Since our dataset is slightly unbalanced, we provide detailed metrics in addition to accuracy, including precision per class, recall, and macro F1 values. We also compare these results to a random baseline.

\subsection{Linguistic Features}
We focus only on content-independent features, in particular syntactic features, following the example of studies on NLI in other languages \citep{malmasi2015norwegian}. Due to the imbalance in the topic distribution in the TLC corpus, we decided not to include lexical features such as word n-grams and contextualized word embeddings in our study. For example, only students with Persian L1 were asked to respond to prompts specifically about happiness and time. This can result in the classifier associating these topics with the languages, rather than discerning the characteristics inherent to the language, thereby introducing a confounding variable to the task \citep{brooke2013native}. By focusing on syntactic features, we aim to capture the underlying syntactic influence of the L1 on its L2 writing independently of the content \citep{malmasi2018portuguese}. We experimented using a combination of two syntactic features: part-of-speech n-grams and function words.

\begin{figure}[ht]
\centering
\begin{tabular}{cccc}
\textit{Oda*} & \textit{yurtta} & \textit{çok} & \textit{kirli.} \\
NOUN & NOUN & ADV & ADJ \\
\end{tabular}

\vspace{10pt} 

\textit{POS 3-gram Example:} (NOUN ADV ADJ) \\
\textit{POS-Lexical 3-gram Example:} (NOUN çok ADJ) \\

\vspace{5pt} 

\caption{An example of a Turkish sentence with a syntactic error characteristic of a Persian L1 learner. The sentence, intended to mean "The room in the dorm is very clean," incorrectly places the modifier (\textit{yurtta}) after the head noun (\textit{Oda}). This is a direct transfer from Persian's head-initial Ezafe construction (\textit{otâq-e dar xâbgâh}). The correct head-final Turkish structure is: \textit{Yurttaki oda çok kirli.}}
\label{fig:pos_example_persian_dorm}
\end{figure}

 \textbf{Function words} are content-independent words, including prepositions, articles, and auxiliary verbs, that play a crucial role in conveying grammatical relationships between words. It is often challenging for L2 speakers to use the appropriate function words and production errors may be due to the influence of their L1 \citep{schneider2016detecting}. These function words are recognised as valuable features for the NLI task. We extracted 120 Turkish words from different grammatical categories including delexicalised verbs. However, it's worth noting that many grammatical aspects, which are morphologically expressed in Turkish, may not be as strongly captured as they would be in English.

\textbf{Part-of-Speech (POS) tags} are linguistic categories or word classes that signify the syntactic role of each word in a sentence. They include basic categories such as verbs, nouns, and adjectives. Assigning POS tags to words in a text introduces a level of linguistic abstraction, meaning that we can work with the underlying structure rather than the content. We use the Turkish NLP toolkit\footnote{\href{https://huggingface.co/turkish-nlp-suite/tr_core_news_trf}{turkish-nlp-suite/tr\_core\_news\_trf}} from SpaCy \citep{honnibal2020spacy} to extract universal POS tags, from which we create n-grams of sizes 1 to 3. These n-grams serve to capture preferences for specific word classes and their localized ordering patterns. Our experiments indicated that sequences of order 4 or higher lead to lower accuracy due to the limited size of our corpus. Therefore, we excluded such higher-order n-grams from our analysis. When both features are combined, function words retain their form instead of being replaced by their POS tags. Figure \ref{fig:pos_example_persian_dorm} provides an example of these extraction steps.

\section{Results}
In this section, we present the results in terms of accuracy achieved by individual feature types. Subsequently, we report the performance obtained using the combination of all features. Finally, we examine the performance obtained by the best system for each L1 class. 

\begin{table}[htbp]
  \centering
  \renewcommand{\arraystretch}{1.2}
  \begin{tabular}{lc} 
    \toprule
    \textbf{Feature Set} & \textbf{Macro F1 (\%)} \\
    \midrule
    Random Baseline & 33.3 \\
    \midrule
    POS 1-grams & 41.9 ($\pm$ 6.6) \\
    POS 2-grams & 46.5 ($\pm$ 7.1) \\
    POS 3-grams & 52.0 ($\pm$ 5.9) \\
    Function Words & 46.7 ($\pm$ 7.0) \\
    \midrule
    \textbf{Full Combination} & \textbf{58.2 ($\pm$ 3.4)} \\
    \bottomrule
  \end{tabular}
  \caption{Average 5-fold cross-validation results. Scores are reported as Macro F1-score percentages with standard deviation.}
  \label{tab:ablation_results}
\end{table}

Table~\ref{tab:ablation_results} displays our main results, comparing the performance of systems trained with different feature sets. All feature types individually outperform the random baseline, demonstrating that both structural and lexical patterns contain useful signals for this task. We observe a clear trend where performance increases with the complexity of the n-grams, with POS 3-grams (52.0\% Macro F1) being the single best-performing syntactic feature set. However, the Full Combination (POS-Lexical) model, which integrates both POS n-grams and function words, achieves the highest overall performance with a Macro F1-score of \textbf{58.2\%}. The notable increase in the average score suggests that the lexical features provide a valuable complementary signal to the purely structural ones.

\begin{table}[htbp]
  \centering
  \renewcommand{\arraystretch}{1.2}
  \begin{tabular}{cccc}
    \toprule
    \textbf{L1} & \textbf{Precision} & \textbf{Recall} & \textbf{F1-score} \\
    \midrule
    Albanian & 0.70 & 0.66 & 0.68 \\
    Arabic & 0.49 & 0.56 & 0.52 \\
    Persian & 0.57 & 0.52 & 0.55 \\
    \midrule
    \textbf{Average} & \textbf{0.59} & \textbf{0.58} & \textbf{0.58} \\
    \bottomrule
  \end{tabular}
  \caption{Aggregate per-class results for the best-performing Full Combination model across all 5 folds. The average scores correspond to the overall performance shown in the aggregate confusion matrix.}
  \label{tab:per_class_results}
\end{table}

\begin{figure}[htbp]
  \centering
  \includegraphics[width=0.45\textwidth]{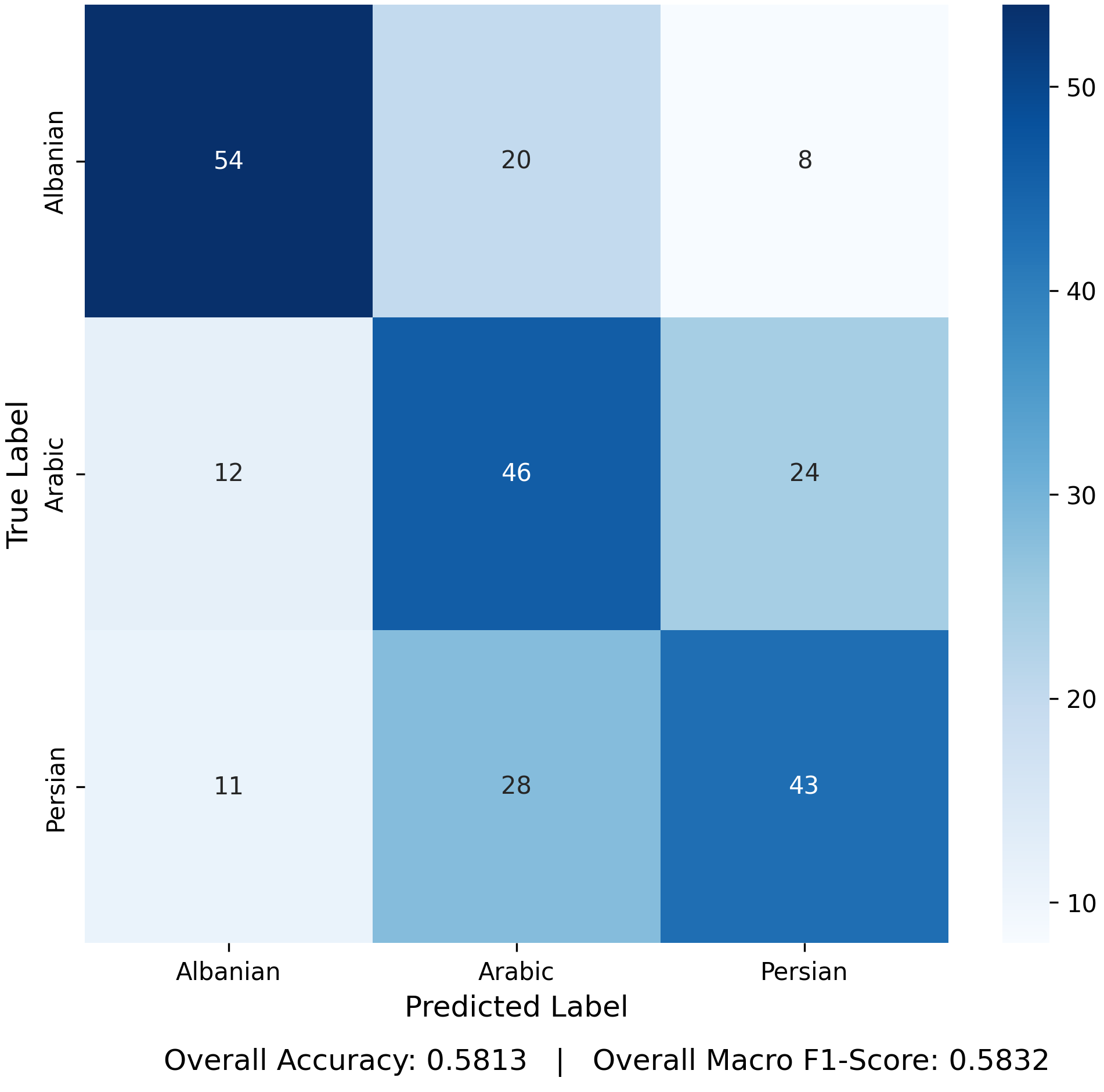} 
  \caption{Aggregate confusion matrix for the best-performing full combination model from the 5-fold cross-validation. The overall Macro F1-score is 58.2\%.}
  \label{fig:confusion}
\end{figure}

Table~\ref{tab:per_class_results} provides a more detailed breakdown of the per-class results for our best-performing Full Combination model, based on the aggregate performance across all five folds. The model achieves its highest performance on Albanian (68\% F1-score), indicating that this L1 group exhibits a highly distinct and consistent set of features in their L2 Turkish. The performance on Arabic (52\% F1) and Persian (55\% F1) is lower, suggesting a greater degree of similarity in the L1 transfer effects from these languages. To provide a visual representation of these findings, we present an aggregate confusion matrix in Figure~\ref{fig:confusion}.

\section{Discussion}
The most significant source of error is the bidirectional confusion between the Arabic and Persian classes. This difficulty appears to be linked to the deep historical and lexical connections between Arabic, Persian, and Turkish. An analysis of the features the model found most predictive (see Appendix, Figures~\ref{fig:appendix_pos_lexical} and \ref{fig:appendix_function_words}) suggests that this confusion may be partly explained by a shared lexical heritage that manifests in the learners' L2 Turkish.

Many of the function words that the model identifies as predictive for these two groups are cognates, originating from Arabic or Persian. For instance, the model highlights the use of words like \textit{bazı} (from Arabic \textit{ba`\d{d}}) and \textit{şey} (from Arabic \textit{\v{s}ay'}) as indicators for Arabic speakers. Similarly, for Persian speakers, top features include conjunctions such as \textit{ama} (from Arabic and Persian \textit{'amm\={a}}), \textit{halbuki} (from Perso-Arabic compound hâl ân ke), and \textit{ve} (from Arabic and Persian \textit{wa}). The frequent use of this shared pool of Perso-Arabic loanwords by both learner groups likely creates lexical ambiguity, making it challenging for the classifier to separate them based on function word choice alone.

However, the model does identify subtle differences in preferred syntactic patterns, which may be attributable to L1 transfer. The feature plots show that Persian L1 writing is strongly associated with n-grams involving the conjunction \textit{ve}, such as \texttt{verb\_ve} and \texttt{ve\_noun\_verb}. This suggests a frequent pattern of explicitly conjoining clauses. A possible explanation lies in the contrast between Turkish and Persian verb chaining. While Turkish heavily utilizes converbs (e.g., the suffix \textit{-ip}) to link consecutive actions within a single complex verb phrase, Persian relies more on simple conjunctions like \textit{va} (the origin of \textit{ve}) to connect clauses. The prevalence of these features suggests that Persian speakers may be transferring this conjunctive strategy into Turkish, using \textit{ve} where a native speaker might use a converb. In contrast, the top syntactic predictors for Arabic speakers, such as \texttt{verb\_verb\_adv}, point to different structural preferences, the cause of which is less clear but may relate to different adverb placement conventions in Arabic.

The model's high performance on Albanian is likely due to the absence of these specific signals, combined with the presence of unique syntactic patterns. The feature analysis indicates that the Albanian L1 profile is not characterized by the same set of Perso-Arabic cognates. More importantly, the model identifies patterns of noun phrase elaboration as positive predictors, including \texttt{adj\_noun\_noun}, \texttt{adj\_noun\_verb}, and even the multi-word modifier pattern \texttt{adv\_adj\_noun}. This may be a direct consequence of a major syntactic difference between the two languages. Albanian uses post-nominal adjectives (the adjective follows the noun, e.g., \textit{një makinë e kuqe}, "a car red"), whereas Turkish uses strictly pre-nominal adjectives (\textit{kırmızı bir araba}, "red a car"). In line with previous research in NLI, we present these interpretations as plausible explanations based on known language transfer patterns. We acknowledge, however, that future work with larger and more diverse corpora is necessary to validate the generalizability of these findings.

\section{Conclusion}

In this study, we presented the first application of NLI to L2 Turkish, focusing on learners with Arabic, Persian, and Albanian L1 backgrounds. Our experiments demonstrated the effectiveness of a hybrid feature set that combines syntactic information (Part-of-Speech n-grams) with key lexical cues (function words), achieving a macro F1-score of 58.2\%. Our analysis revealed that while the model could reliably distinguish Albanian learners, its primary challenge was the confusion between Arabic and Persian speakers. We provided evidence suggesting this is due to a shared inventory of Perso-Arabic loanwords, while also identifying distinct syntactic patterns that offer a path to better discrimination.

We identify several promising directions for future research. Firstly, expanding the corpus with more texts and additional L1 groups would be invaluable for building more robust and generalizable models. Secondly, incorporating a a standardized learner proficiency metric as a variable could help disentangle the effects of language transfer from developmental errors. Finally, to better distinguish between linguistically close L1s like Arabic and Persian, future models could incorporate features less susceptible to lexical overlap, such as dependency parse subtrees. For a morphologically rich language like Turkish in particular, a systematic analysis of morphological errors presents a promising future direction for discovering robust features that are highly indicative of L1 influence.

\section*{Limitations}
Our analysis brings attention to two potential limitations. Firstly, the size of our corpus, while sufficient for achieving results significantly above the baseline, is relatively limited compared to large-scale NLI studies for Portuguese and Norwegian \citep{malmasi2018portuguese, malmasi2015norwegian}. It is, however, comparable in scale to the corpus used for the first Arabic NLI study \citep{malmasi2014arabic}. Secondly, we acknowledge that our parser might not be entirely suitable for learner language, which could introduce additional noise into the feature space \cite{van2002effect}. However, we chose to use the spaCy transformer-based model, as a recent comparative study has shown it to be relatively robust for processing learner language compared to alternatives \citep{kyle2024evaluating}, which helps mitigate this concern.

\section*{Ethics Statement}
Our study only processes information from publicly available learner corpora. We place great emphasis on protecting the privacy of individuals and ensure that no sensitive personal data is accessed, stored or processed at any stage of the project. Our research adheres to the ethical guidelines of University of Zurich.

\section*{Acknowledgements}
Partial support for this research was provided by the Swiss Innovation Agency (Innosuisse) under grant number 103.188 IP-ICT (Project: AUCH), a collaboration between the University of Zurich and PRODAFT.

\clearpage
\onecolumn
\begin{appendices}
\section{Feature Analysis Plots}

This appendix visualizes the most discriminative features for each L1 background, as determined by the linear SVM coefficients from our one-vs-rest classification models.

\begin{figure}[htbp!]
  \centering
  \includegraphics[width=0.9\textwidth]{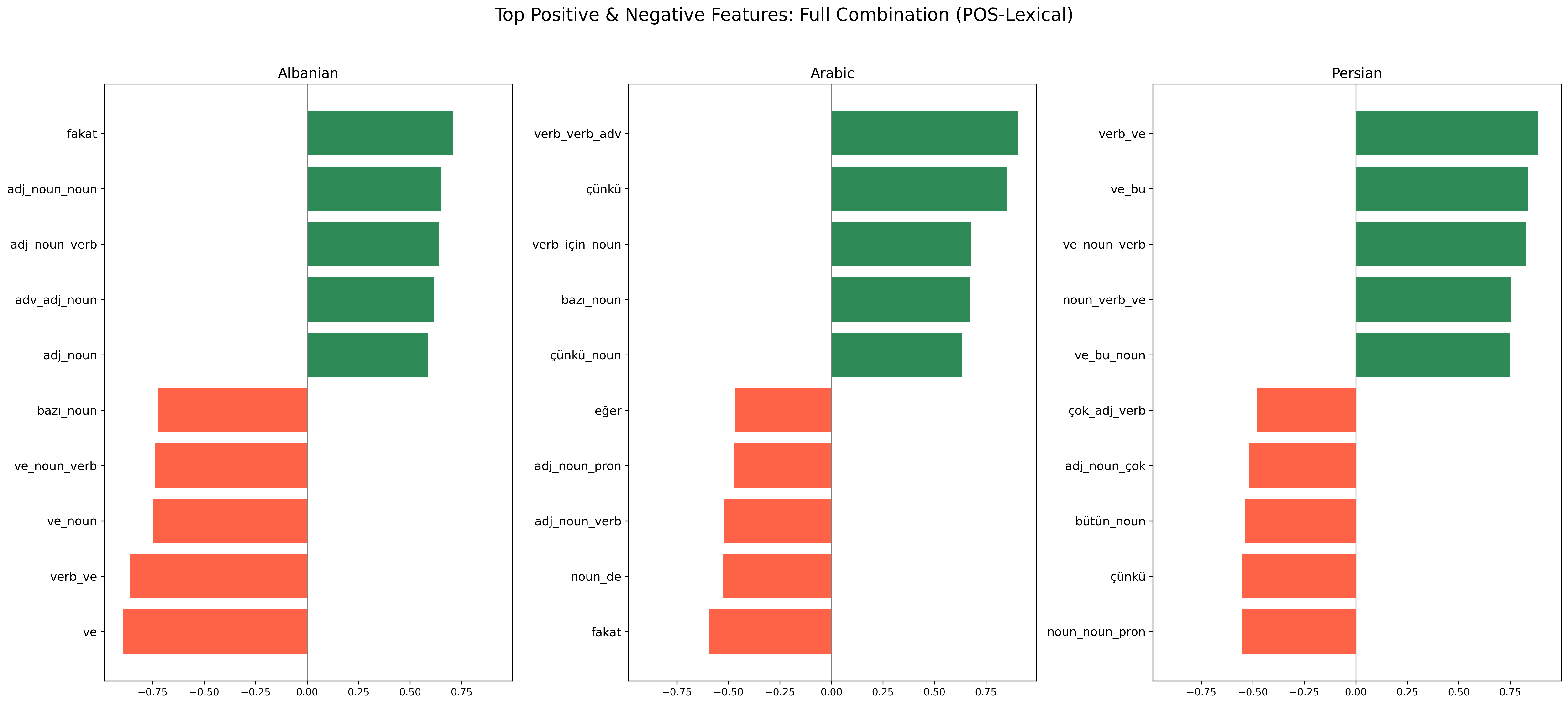}
  \caption{Top features from the Full Combination (POS-Lexical) model, visualizing the syntactic and lexical patterns most predictive for each class.}
  \label{fig:appendix_pos_lexical}
\end{figure}

\begin{figure}[htbp!]
  \centering
  \includegraphics[width=0.9\textwidth]{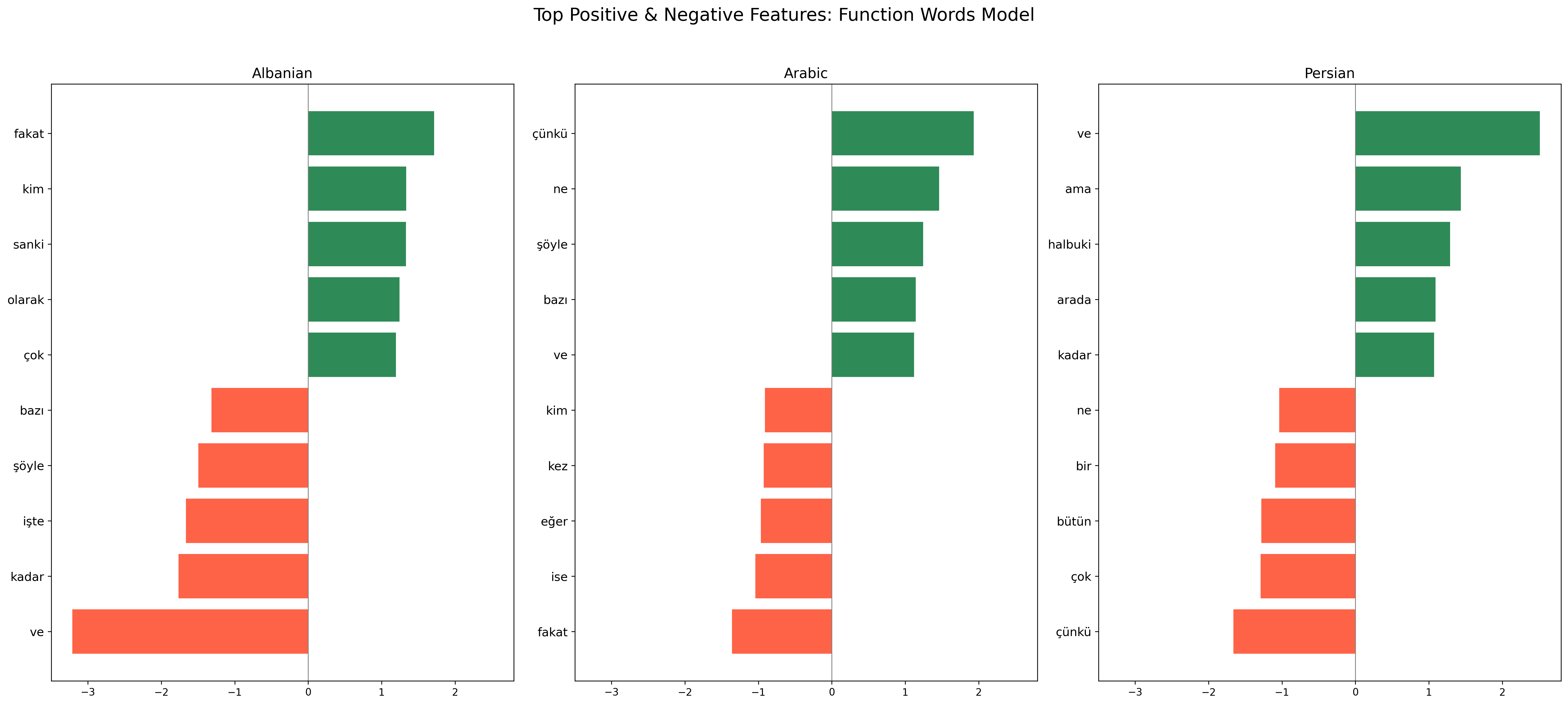}
  \caption{Top features from the standalone Function Words model, visualizing the function words most strongly associated with each L1 group.}
  \label{fig:appendix_function_words}
\end{figure}

\end{appendices}

\clearpage
\twocolumn
\bibliography{anthology,custom}
\bibliographystyle{acl_natbib}

\end{document}